# An AI-Assisted Skincare Routine Recommendation System in XR

Gowravi Malalur Rajegowda[1], Yannis Spyridis[2], Barbara Villarini[3] and Vasileios Argyriou[1]

[1] Dept. of Network ands and Digital Media, Kingston University, London, UK
[2] Dept. of Electronic and Electrical Engineering, The University of Sheffield, Sheffield, UK
[3] School of Computer Science and Engineering, University of Westminster, London, UK

**Abstract.** In recent years, there has been an increasing interest in the use of artificial intelligence (AI) and extended reality (XR) in the beauty industry. In this paper, we present an AI-assisted skin care recommendation system integrated into an XR platform. The system uses a convolutional neural network (CNN) to analyse an individual's skin type and recommend personalised skin care products in an immersive and interactive manner. Our methodology involves collecting data from individuals through a questionnaire and conducting skin analysis using a provided facial image in an immersive environment. This data is then used to train the CNN model, which recognises the skin type and existing issues and allows the recommendation engine to suggest personalised skin care products. We evaluate our system in terms of the accuracy of the CNN model, which achieves an average score of 93% in correctly classifying existing skin issues. Being integrated into an XR system, this approach has the potential to significantly enhance the beauty industry by providing immersive and engaging experiences to users, leading to more efficient and consistent skincare routines.

**Keywords:** Artificial Intelligence, CNN, Deep Learning, Skincare, Recommendation System, VR/AR.

## 1    Introduction

Skin care has been an important aspect of personal hygiene and beauty for centuries, and often involves several steps in order to maintain and improve the texture of the skin. With the advent of new technologies in the medical and pharmaceutical industries, there has been an increase in the number of skin care products available in the market. However, choosing the right skin care product that suits an individual's skin type and concerns can be overwhelming, especially with the ever-increasing number of products available. In recent years, the use of artificial intelligence (AI) in skin care recommendation systems has gained popularity as it helps to personalise skin care recommendations based on an individual's skin type, concerns, and lifestyle.

The motivation behind developing an AI-assisted skin care recommendation system arises from the need to help individuals choose the right skin care products that are suitable for their skin type, taking into consideration their personal needs, while avoiding visits to medical experts in case there are non-clinical skin concerns. The use of AI can help to reduce the guesswork and time spent on researching and selecting skin care products, which can be often time-consuming and inefficient. By using an AI-assisted skin care recommendation system, individuals can get personalised recommendations based on their own needs and treat mild abnormalities in an efficient way.

In recent years, there has been an increase in research on recommendation systems using AI. Several studies have shown that AI models can provide accurate and personalised recommendations based on an individual's needs [1]. For instance, a study conducted by Kumar et al. [2] proposed an AI-assisted college recommendation system that provides personalised recommendations after building a user profile based on certain questions. The system then maps this profile to college profiles scraped from the web based on a content-based approach. Skin recognition algorithms have also been widely used in recent years to help in medical applications. These systems use machine learning models to analyse an individual's skin type and abnormalities and classify it to certain categories. Using microscopic images, Saidah et al. [3] developed a Convolutional Neural Network (CNN) to classify the given image into normal, dry, oily, or combined skin, with a very high accuracy. While there is a lack of extensive research on extended reality (XR) recommendation systems, there are a few studies that aim to provide personalised recommendations to users in an immersive and interactive setting. For instance, Lin et al. [4] proposed a recommender system that employs virtual reality, which serves as a platform for retrieving historical interior design drawings from a database and recommending a prototype drawing to the consigner. Such a system can be quite useful for designers as it enables them to store historical drawing items, extract pertinent design features, and guide consigners from the query system to the historical database, thus accessing most appropriate design drawing that match their interests and requirements.

In this paper, we present a skin care recommendation system, which utilises deep learning models to analyse the skin type of users, and along with targeted inputs from specialised questions, provides specific suggestions with respect to skin care products. Furthermore, a dataset for skin analysis and classification is part of the contributions of this work. To make the skin care recommendation experience more immersive and interactive, we have integrated the AI-assisted skin care recommendation system into an XR platform. By incorporating XR technology into the AI-assisted skin care recommendation system, users can experience a personalised skin care journey, allowing them to see the effects of the recommended products on their skin over time. This can lead to a more engaging experience that encourages individuals to adhere to their personalised skin care routine.

The rest of the paper is organised as follows. Section 2 discusses the related literature on skin issue detection and recommendation systems. Section 3 presents the methodology that was followed for the dataset creation and the training process of the CNN model, while section 4 discusses the system evaluation. Finally, section 5 concludes the paper.

## 2 Literature Review

### 2.1 Skincare Issues and Importance of Product Ingredients

In a survey conducted by El-Essawi et al. [5], it was found that a large group of individuals in the USA suffer from multiple skin issues. Table 1 illustrates the distribution of the most common skin concerns identified.

**Table 1.** Distribution of common skin issues

| Skin issue | Percentage of population in the sample (%) |
|---|---|
| Uneven skin tone | 56.4 |
| Skin discoloration | 55.9 |
| Dry skin | 51.9 |
| Acne | 49.4 |
| Wrinkles | 39.4 |
| Moles | 32.4 |
| Rashes | 30.9 |
| Oily skin | 23.2 |

It is observed that roughly 55% of the participants suffer from conditions such as uneven skin tone, discolouration, and about 50% suffer from acne or dry skin. Other issues such as wrinkles, redness, rashes, and oily skin are also quite common, while it should be noted that most participants suffer from multiple skin issues. To solve these conditions, one of the proven scientific methods is the adaption to a carefully curated skin routine. However, identifying such a routine can be complex, because it involves relying upon multiple skin care products, examining their ingredients, their proportion, and impact [5].

In a study by Rodan et al [6] it was found that daily skincare routines may have statistically significant long-term effects on the overall quality of one's skin health and complexion. Basic skin care needs involve processes of skin protection, issue prevention, cleansing, and moisturising. Ingredients such as either zinc oxide or avobenzone for example can help block UV-B and UV-A effects on the skin. While the skincare routine is complex, cosmetic stores still recommend products based on popularity with less regard to each individual's skin conditions and issues they are seeking to solve [7]. Since active ingredients in the product and their knowledge are hard to procure due to the vast assortment of alternatives, it is important to enable a solution that assists users in recommending products based on ingredients and alternatives and to personalise this recommendation based on their skin issues and conditions.

### 2.2 Skin Detection Models

Arifin et al. [8] developed an automated dermatological diagnostic system that detects and identifies skin anomalies using high-resolution color images and patient history. The system uses color image processing techniques, clustering, and neural networks to achieve high accuracy rates of 95.99% for detecting diseased skin and 94.016% for identifying diseases. ALEnezi [9] proposed an image processing-based method to detect skin diseases of the affected skin area in provided images. The approach works on the inputs of a color image and resizes the image to extract features using a pretrained CNN. The system then classifies the features using a Multiclass SVM and shows the results to the user, including the type of disease, spread, and severity.

With respect to non-clinical skin issues, Alamdari et al. [10] compared image segmentation methods to detect acne lesions and several machine learning methods to distinguish between different types of such lesions. Two-level k-means clustering was found to outperform other techniques with an accuracy of approximately 70% for detecting acne lesions. The accuracy of differentiating between acne scarring and active inflammatory lesions was 80% and 66.6% for fuzzy-c-means and the SVM method, respectively. The performance accuracy of classifying the normal skin from detected acne lesions was 100% using fuzzy-c-means clustering.

### 2.3 Skincare Recommendation Systems

A study by Hsia et al. [11] utilised machine learning to process and classify multiple features of skin quality and acne status to provide recommendations for facial skincare products. The proposed system, which relies on an SVM classifier was evaluated on 15 subjects and resulted in a consumer satisfaction index of 80%.

Similarly, Lin et al. [12] proposed a new business model of facial skincare products that utilises computer vision technology. The framework consists of a finger vein identification system, a skincare product recommendation system, and an electronic payment system. Experimental results showed that the finger vein identification system had the lowest equal error rate and shortest response time, while the skin type classification accuracy was the highest. Lee [7] proposed a skincare product recommendation system that uses content-based filtering to suggest products based on a user's skin type and desired beauty effect. The system analyses the chemical composition of products to find those with similar ingredient compositions.

## 3 Methodology

The implementation workflow of our AI-assisted skincare recommendation system is presented in Fig.1. The system first relies on a two-stage process to analyse provided facial images, extract relevant features, and identify existing skin issues. The outcome of this process is then fed to the skincare routine recommendation algorithm, which considering the similarities between ingredients in the product catalog, suggests a series of products that target the specific needs identified in the previous process.

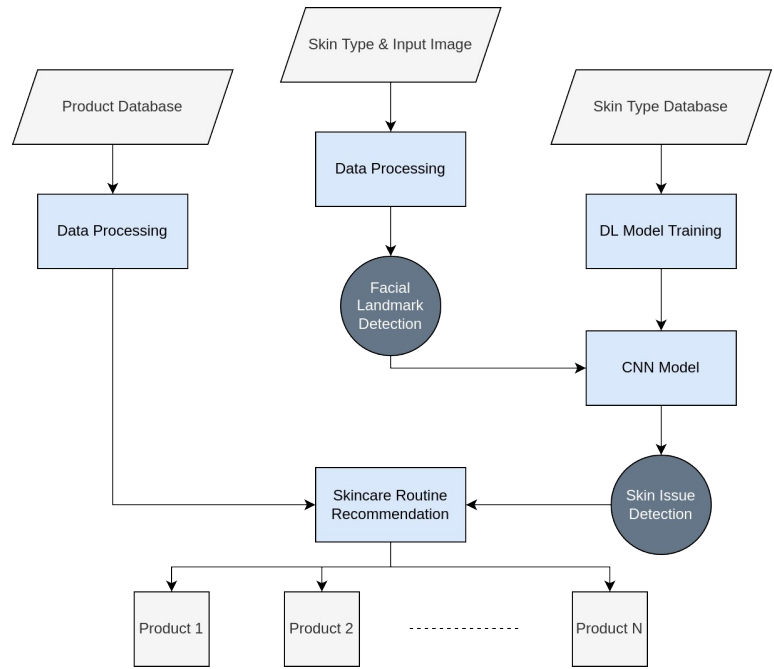

**Fig. 1.** Implementation workflow

### 3.1 Facial Landmark Detection Subsystem

This subsystem is used to detect important characteristics in the provided facial images, such as eyes, forehead, cheeks, and chin, that are considered predominant areas in which non-clinical skin issues can be observed. The developed algorithm relies on the "Haar Cascade" classifier, which is an object detection algorithm commonly used in face detection applications, using a machine learning approach to identify objects in image streams. The algorithm is based on the "Haar Wavelet" [13] function, which is used to extract key features from an image, then combine them to detect objects of interest.

More specifically the system utilises, the "Eye Cascade", a specific "Haar Cascade" classifier, designed to detect eyes in the provided image. This classifier is trained on a set of positive and negative images and learns to differentiate between the ones that contain eyes, thus extracting features that are important for eye detection. The location of the rest of the facial characteristics is determined using the "Shape Predictor 68 Face Landmark", which is a machine learning model, capable of predicting the locations of 68 specific facial landmarks on a human face.

Following the prediction of the eye detection and the facial landmark models, four main regions of the face are spliced and stored to be provided as input to the Skin Issue Recognition subsystem. This process is implemented using the "OpenFace" face recognition library, which relies on a combination of convolutional neural networks (CNN) and recurrent neural networks (RNN) to extract the facial features from the given images.

### 3.2 Skin Issue Recognition Subsystem

**Dataset Collection and Data Preprocessing**

Due to the lack of large publicly available datasets for skin issues, a new dataset was developed by collecting free images from different sources on the Internet. The dataset contains approximately 3500 images of four labeled classes: a) Acne, b) Pigmentation, c) Wrinkles, and d) Clear Skin. The images were processed so that the background noise was removed, and the exposure was adjusted to normal values to ensure robust training.

The final images were subjected to the Facial Landmark Detection system presented in subsection 3.1, so that the relevant skin patches from each image is extracted and saved along with the correct label. Finally, a process of data augmentation was followed whereby a data generator was utilised to apply image augmentation options such as zoom, shear, flip, and brightness adjustment. Through the augmentation, the amount and diversity of data in the training dataset was increased, resulting in a model more robust to variations in the input data, and capable of generalising to a wider range of input conditions.

**Training Process**

For the training process, we utilised transfer learning using the VGG16 model, due its relatively small number of parameters, consistent block structure, and modular nature, which allows it to be easily adapted for other tasks. The VGG16 architecture consists of 16 layers of convolutional and pooling layers, followed by three fully connected layers for classification. The convolutional layers use small filters (3x3) with a stride of 1 pixel and padding to preserve spatial resolution, while the pooling layers use max pooling with a 2x2 filter and a stride of 2 pixels to reduce the spatial dimensions.

After splitting the dataset into 80%, 15% and 5% for training, testing, and validation respectively, we trained the model ustilising the VGG16 pretrained weights, to classify the input image into one of the four labels. We used SGD-momentum as the optimiser which is an extension of the gradient descent algorithm. In SGD-momentum, a momentum term is introduced that accumulates the gradient over previous iterations and adds it to the current update, to smooth out the updates and reduce oscillations in the parameter space. The update for a parameter $\theta$ at iteration $t+1$ is given by:

$$\theta_{t+1} = \theta_t - V_t, \tag{1}$$

$$V_t = m * V_{t-1} + lr * gradient, \tag{2}$$

where $V_t$ is the momentum vector at iteration $t$, $m$ is a hyperparameter that controls the contribution of the previous velocity to the current update, and $lr$ is the learning rate. Using the momentum update helped improve the accuracy and solve over-fitting problems, while allowing for a higher learning rate. Ultimately, this accelerated the convergence without destabilising the optimisation process, as demonstrated in Fig. 2. The training and validation losses throughout the training process are shown in Fig. 3.

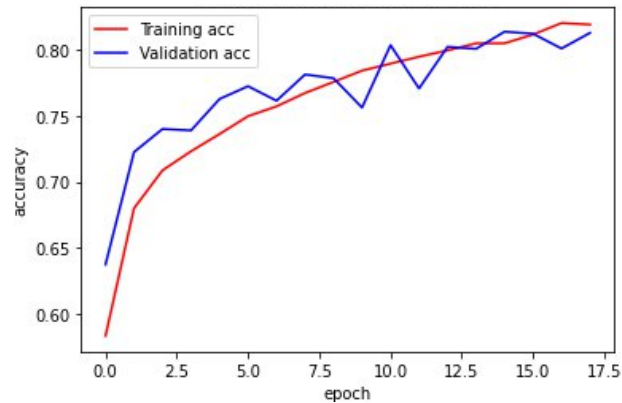

**Fig. 2.** Training and validation accuracies.

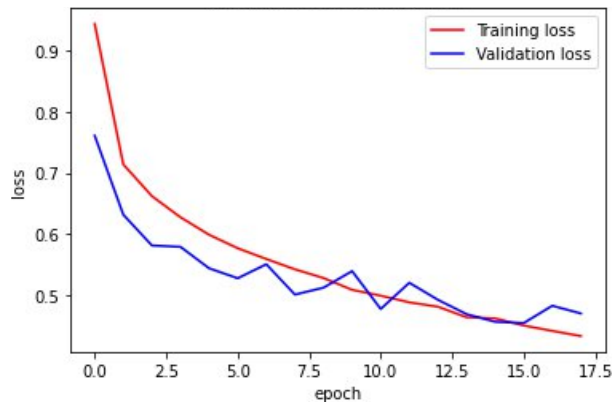

**Fig. 3.** Training and validation losses.

### 3.3    Product Recommendation Engine

The skincare recommendation system presented in this paper utilises a content-based approach to product recommendations, specifically focusing on ingredient similarity within the same product category. This method distinguishes itself from other recommendation systems, as skincare requires personalised care and attention based on individual skin types. As a result, a review-based recommendation system may not be appropriate in this context and therefore the developed recommendation system is designed based on the content-based approach necessitated by the complexity of skincare. By leveraging ingredient information and comparing products within the same category, we provide a more personalised and accurate recommendation for users. The presented approach accounts for the unique needs and characteristics of each user's skin type, as designated by the Skin Issue Recognition subsystem, resulting in a more effective and efficient skincare recommendation process.

**Dataset Preparation**

For the skincare product recommendation engine, we utilised an existing dataset [14] that focuses on day-to-day skincare routines. The primary categories considered in this dataset include cleanser, serum, treatment, moisturizer, and sunscreen products. The data was sourced from a popular skincare website and included information such as ingredients, price, color, brand, and chemical components. The dataset consists of 1,472 unique items and 17 columns. The utilisation of this dataset allowed us to focus the analysis on the most relevant skincare categories and ensured that the results were based on a comprehensive and representative set of data. A sample of the dataset is presented in Table 2.

**Table 2.** Sample of the skincare product dataset.

| ID | Label | Issue | Brand | Name | Ingr. | Combin | Dry | Oily |
|----|-------|-------|-------|------|-------|--------|-----|------|
| 1 | Moistu.. | Acne | LA MER | Crème… | Algae,… | 1 | 1 | 1 |
| 2 | Moistu.. | Acne | SK-II | Facial… | Galact.. | 1 | 1 | 1 |
| 6 | Moistu.. | Acne | DRUNK.. | Protin… | Dicapr… | 1 | 1 | 1 |
| 11 | Moistu.. | Acne | BELIF | The Tru. | Diprop… | 1 | 0 | 1 |

The dataset was adjusted with the specific goal to address the challenges associated with predicting skincare products based on past usage in mind. Reliance solely on past usage to make recommendations can be unreliable and unpredictable due to the vast domain of skincare and the personalised nature of cosmetic recommendations. Instead, the developed recommendation system focuses on identifying the ingredients in each product and comparing them to identify similarities between products. Preprocessing steps were conducted to prepare the dataset for analysis, including removing duplicate items and cleaning the text data in the "Ingredient" column. Additional information, such as compositions and ingredients, was also collected to ensure that the final dataset was comprehensive and representative.

The analysis focused on products that were classified based on different skin types and skin concerns to provide personalised recommendations. Specifically, the main dataset includes five recommended products in each of the five categories: cleanser, moisturizer, treatment, mask, and sunscreen. By mapping these categories to specific skin types, such as "Cleanser" for "Dry Skin," targeted recommendations are provided that are customised to each user's individual needs and preferences. The content-based approach which focuses on ingredient similarity between products, allows for reliable and accurate recommendations despite the personalised nature of skincare.

**Determining Similarity of Products**

To determine the similarity of ingredients between products, the t-SNE technique is employed, leading to the reduction of the dimensionality in the data. By preserving the similarities between instances, t-SNE effectively visualizes high-dimensional data on a two-dimensional plane. Similarities are calculated based on the distances between data points, and cosine similarity is used to find similarities between non-zero vectors. Unlike distance-based measures, Cosine Similarity captures more information about vector direction.

In the developed system, this technique is applied when the user selects a known brand on the recommender system. The system then analyses ingredient similarities based on skin type and skin concern and recommends a complete skincare routine with up to five products in each category, based on t-SNE and Cosine Similarity.

**Recommendation through Matrix Factorisation**

The skincare industry can be overwhelming for users due to the vast number of products available. To simplify the process, the developed recommendation system has been designed to consider user input before suggesting products. The system considers two inputs - brand name or desired product and the skin concern detected by the CNN model of the Skin Issue Recognition subsystem - using the Matrix Factorisation method. This method is ideal as it is non-biased towards sparse data.

The Matrix Factorisation method calculates two factors: user input features and product similarity based on predicted skin issues from the products dataset. The system then reduces dimensions based on skin issues, brands, skin type, and ingredients to provide recommendations. For instance, if the user has acne problems, a product that suits their needs is recommended based on the comparison of the similarity scores among relevant products. Therefore, the system suggests 5 products that are nearest to the selected product based on their ingredients. The recommendation process contains a few different steps, which are followed in an XR setting, as detailed in the following subsection.

**XR Integration**

The XR platform was developed to offer an immersive experience to users when going through the skincare recommendation process. The platform offers an interface of widgets through which the application communicates with the deployed AI models, so that the individual can get a personalised recommendation for their skincare routine. The first such widget allows users to upload their facial image, which is provided as input to the Facial Landmark and Skin Issue Recognition subsystems. The recognised faces are processed and uploaded in a queue (Fig.4). The faces are analysed by the Deep Neural Network, which provides the skin classification. Using this information, the Recommendation Engine provides the suggestions for the skincare routine as a list of products. In the final widget, the user can select alternatives brands, and the Recommendation Engine suggests corresponding products based on the similarity of the ingredients identified. The system also keeps track of uploaded images, allowing users to view the results of the followed skincare routine over time.

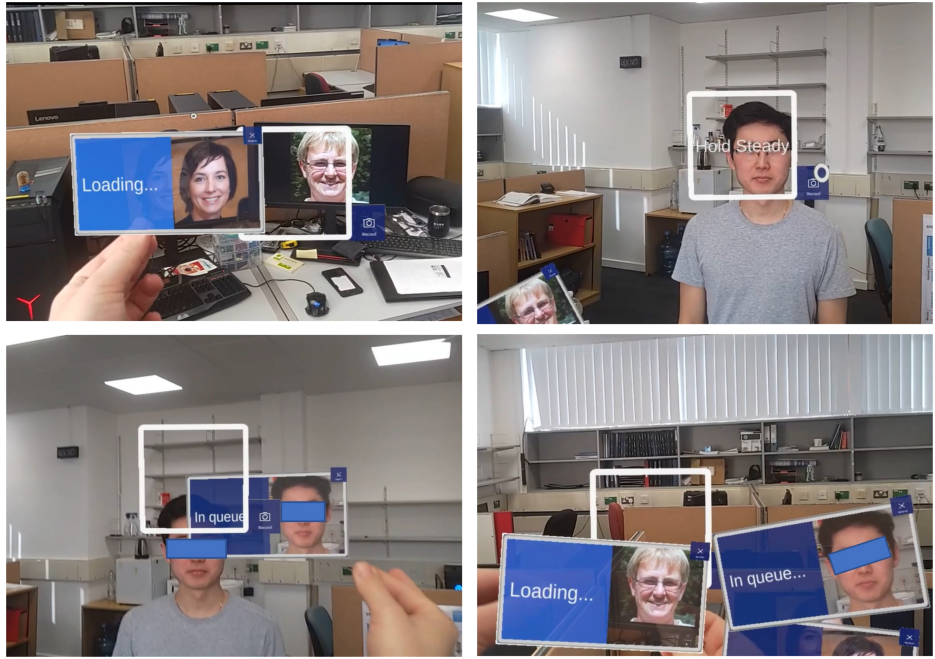

**Fig. 4.** The automatic recognised faces are uploaded in the queue ready to be analysed by the deep neural network, which provides the skin classification.

## 4 System Evaluation

### 4.1 Facial Landmark Performance

The system's performance in detecting skin patches was evaluated under varying lighting conditions and was found to be effective in detecting the eyes and mouth in the uploaded image. The accuracy of the skin patches was assessed using confusion metrics, which compared the classification results of already classified images to the live classification. The results of the analysis are presented in the confusion matrix of Fig.5.

|  | Right cheek | Left cheek | Chin | Forehead |
|---|---|---|---|---|
| Right cheek | 0.98 | 0 | 0 | 0 |
| Left cheek | 0 | 0.98 | 0 | 0 |
| Chin | 0 | 0 | 0.96 | 0 |
| Forehead | 0 | 0 | 0 | 0.95 |
| Not detected | 0.02 | 0.02 | 0.04 | 0.05 |

**Fig. 5.** Face landmark detection confusion matrix.

The findings reveal that the face landmark detection system was highly accurate, with a success rate of over 95% in correctly categorising the spliced area based on facial landmarks for all cases. However, there were a few instances in which the system failed to correctly separate the face landmarks, but these were due to the forehead, chin, or cheek not being fully visible in the image. This issue was slightly more pronounced for the chin, particularly when the person was not facing straight. Despite this minor limitation, the overall recommendation system designed in this work was not significantly impacted by this issue. Nevertheless, based on these results users are advised to show their full face as much as possible during usage, to minimise the occurrence of this issue.

### 4.2 Skin Issue Recognition Performance

The evaluation of the CNN model utilised by the Skin Issue Recognition subsystem involved the use of several metrics, such as a) Validation accuracy, b) Precision, c) Recall, and d) F1-score. The validation accuracy was determined during the training process by computing the ratio of correctly classified images to the total number of guesses made by the model. Precision and recall are metrics that are computed independently of validation accuracy, and their scores are determined after the model is trained with an unbiased set of data. Precision is a measure of the accuracy of a model's positive predictions, while recall measures the actual positives of the model. The formulas for precision and recall are as follows:

$$Precision = \frac{True\ Positives}{True\ Positives + False\ Positives},$$

(3)

$$Recall = \frac{True\ Positives}{True\ Positives + False\ Negatives} \tag{4}$$

F1 score is a function of both recall and precision and is computed as an average weighted by precision and recall. This is especially important when there is an uneven class distribution, which is the case in our model. The formula for F1 score is as follows:

$$F1 = 2 * \frac{Precision * Recall}{Precision + Recall} \tag{5}$$

The confusion matrix of the success rate is shown in Fig. 6, while the evaluation of the above metrics in our modified VGG16 model are presented in Table 3.

As observed in the table, the model achieved a very high accuracy across all classes, achieving the top performance in the "Acne" case with a score of 96%, and in general being capable of correctly classifying most of the samples in the dataset. However, due to the imbalanced structure of the dataset, it is important to examine the rest of the metrics to get a more complete understanding of the performance. When investigating the Precision, we still observe a high percentage of correctly identified positive samples as true positives, except for the "Pigmentation" case.

**Table 3.** Precision, recall, and F1-score metrics.

| Skin issue | Precision | Recall | F1-score | Accuracy |
|---|---|---|---|---|
| Acne | 0.83 | 0.76 | 0.79 | 0.96 |
| Clear Skin | 0.84 | 0.91 | 0.87 | 0.94 |
| Pigmentation | 0.65 | 0.77 | 0.71 | 0.91 |
| Wrinkles | 0.88 | 0.52 | 0.66 | 0.89 |

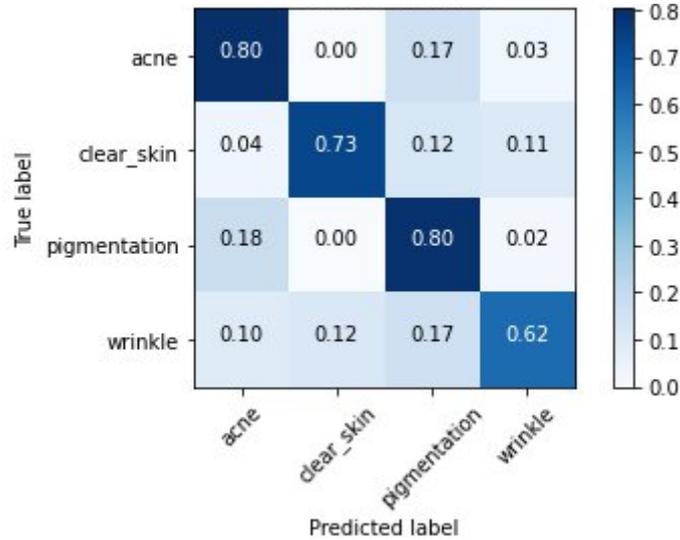

**Fig. 6.** Skin issue recognition confusion matrix.

A similar trend is observed in the Recall metric, but in this case the low score affects the "Wrinkles" class, indicating the difficulty of the model in identifying true positive samples in this category. These metrics also reflect the performance in F1-score, which is the harmonic mean of Precision and Recall and in our case depict a balanced performance in the "Acne" and "Clear Skin" classes, but a lower score in the other two classes.

## 5    Conclusion

Skincare is a crucial aspect of personal care that enables individuals to maintain healthy skin. This practice typically involves the use of products prescribed by dermatologists, cosmetologists, or sourced from various online platforms. Skincare encompasses a wide range of topics that require extensive research, such as ingredients, skin types, and skin concerns in order to identify the appropriate products for one's use case. As the skincare industry continues to grow, research on cosmetics has shown significant impacts on the overall appearance and health of the skin. With the rise of the global pandemic, individuals have increasingly invested more time and financial resources in self-care, turning to online sources for guidance in a process that is often inefficient.

Skincare products contain varying ingredients, and selecting the appropriate ones for different skin types and concerns can be challenging and complex. This paper aimed to enhance and automate the process of diagnosing these concerns, analysing skin types and finally offering suitable product recommendations tailored to an individual's skincare needs. Towards this goal, a comprehensive recommendation system was designed and developed, including a Facial Landmark Detection subsystem, a Skin Issue Recognition algorithm, and finally the Skincare Product Recommendation engine. Key focus was especially given in the CNN model that was built to analyse the users' skin type. The model relies on a modified version of the VGG16 architecture, which in combination with the Facial Landmark Detection subsystem is able to recognise and classify the skin type into one of four classes. The model's performance is highly adequate for this use case, achieving an overall average accuracy of 93%. This output is then used by the Skincare Recommendation engine along with related user questions to recommend a series of products that target the specific needs of the individual.

Through the integrated platform, users are able to employ a one-stop solution for treating their skin issues, by just using an image and providing minimal input to the system, which then automatically recommends an appropriate skincare routine. By incorporating the process in an XR environment, users can interact with the system in an immersive and engaging manner, that plays a vital role in identifying and adhering to the routine consistently. While the system displays high potential in the skincare domain, there are still limitations with respect to the performance of the Skin Issue Recognition algorithm. More specifically, as suggested by the precision and recall metrics, there is room for improvement in identifying tricky skin issues, such as pigmentation or wrinkles. In addition, the current system supports four categories of skin issues, therefore limiting the extent of the target audience. The above issues could be solved by employing a more extensive dataset, that includes additional classes and poses, further fine-tuning and retraining of the CNN model.